\documentclass[conference]{IEEEtran}
\IEEEoverridecommandlockouts
\usepackage{cite}
\usepackage{amsmath,amssymb,amsfonts}
\usepackage{algorithmic}
\usepackage{graphicx}
\usepackage{textcomp}
\usepackage{xcolor}
\usepackage{subfig}
\usepackage{hyperref}

\def\BibTeX{{\rm B\kern-.05em{\sc i\kern-.025em b}\kern-.08em
    T\kern-.1667em\lower.7ex\hbox{E}\kern-.125emX}}
    
\begin{document}

\title{Wind speed prediction using multidimensional convolutional neural networks}

\author{\IEEEauthorblockN{Kevin Trebing}
\IEEEauthorblockA{\textit{Department of Knowledge Engineering} \\
\textit{Maastricht University}\\
Maastricht, The Netherlands \\
k.trebing@student.maastrichtuniversity.nl}
\and
\IEEEauthorblockN{Siamak Mehrkanoon*\thanks{*corresponding author.}}
\IEEEauthorblockA{\textit{Department of Knowledge Engineering} \\
\textit{Maastricht University}\\
Maastricht, The Netherlands \\
siamak.mehrkanoon@maastrichtuniversity.nl}
}

\maketitle

\begin{abstract}
Accurate wind speed forecasting is of great importance for many economic, business and management sectors. This paper introduces a new model based on convolutional neural networks (CNNs) for wind speed prediction tasks. In particular, we show that compared to classical CNN-based models, the proposed model is able to better characterise the spatio-temporal evolution of the wind data by learning the underlying complex input-output relationships from multiple dimensions (views) of the input data. The proposed model exploits the spatio-temporal multivariate multidimensional historical weather data for learning new representations used for wind forecasting. We conduct experiments on two real-life weather datasets. The datasets are measurements from cities in Denmark and in the Netherlands. The proposed model is compared with traditional 2- and 3-dimensional CNN models, a 2D-CNN model with an attention layer and a 2D-CNN model equipped with upscaling and depthwise separable convolutions.
\end{abstract}


\section{Introduction}
Weather conditions undeniably affect many aspects of life in modern societies. Adverse weather conditions and events can have both direct and indirect effects on a large number of economic and business sectors, such as transport, logistics and agriculture. Accurate and timely weather forecasts are therefore important for a variety of applications by facilitating planning and management in response to weather conditions. Weather forecasts are crucial to predict natural disasters such as extreme rainfalls, floods, hurricanes and heat waves. With wind power gaining importance as a source of renewable energy in recent years, wind speed forecasting has become an important tool for efficient and adaptive management of wind parks \cite{li2010comparing}.

To date, weather forecasting typically relies on numerical weather prediction (NWP) models to solve complex mathematical equations that simulate the real world's atmosphere, fluid- and thermodynamics as closely as possible \cite{soman2010review}. This approach requires immense computing power and, even with current technological equipment and tools, can take several hours to process \cite{agrawal2019machine}. Due to their extensive computational requirements, the applicability of NWP models is, in practice, restricted to more long-term predictions. That is, NWP models can be used to predict for instance three hours ahead, however, as the time to process this prediction takes longer than three hours, the practical utility of this approach becomes limited \cite{agrawal2019machine}. This results in an important gap for short-term weather predictions that could enable short-term planning. 

One way this gap has been addressed so far is through data-driven machine learning models that have been able to significantly reduce the processing time compared to NWP models (e.g. \cite{agrawal2019machine, sonderby2020metnet}). Machine learning-based models have been successfully applied in various application domains \cite{mehrkanoon2014parameter, mehrkanoon2019cross,mehrkanoon2018deep,mehrkanoon2017regularized}. In particular, the literature has witnessed the use of different classes of machine learning models for time-series prediction tasks. They range from kernel-based models such as support vector machines \cite{sapankevych2009time}, Least squares support vector machines \cite{van2001financial} as well as deep artificial neural networks-based models among others \cite{scher2018toward, hossain2015forecasting,rodrigues2018deepdownscale}. Machine learning-based models for weather forecasting can potentially reduce the need for large computing power relative to the requirements of current NWP models. A crucial difference between the two approaches to weather forecasting lies in what they base their predictions on. Unlike conventional NWP models, data-driven approaches often do not make any assumptions about the underlying physical factors influencing weather. Instead, they use historical weather observations to train a machine learning model to map the exogenous input to a target output \cite{mehrkanoon2019deep}.  
 
In recent years, convolutional neural networks (CNNs) and their variants have emerged as a powerful computational architecture for addressing a range of challenging tasks \cite{krizhevsky2012imagenet,mehrkanoon2019deepneural}. Convolutions are kernel-based operations that employ a sliding window approach over a matrix or tensor. The use of kernels allows to capture local invariant features and utilising weight sharing decreases the number of trainable parameters of the network significantly \cite{krizhevsky2012imagenet}.

The advancement of modern deep learning techniques has inspired many researchers to explore the available massive amount of weather data. The prediction performance of deep learning architectures for the purpose of weather forecasting have been studied in \cite{salman2015weather}. The authors in \cite{grover2015deep} combined predictive models with a deep neural network to model the joint statistics of a set of weather-related variables.
In this work, we propose a novel convolutional neural network architecture to model the underlying complex nonlinear dynamics of wind speed in multiple Danish and Dutch cities over the course of up to 24 hours. This is achieved by reshaping the data in a way that allows the use of convolutions over multiple input dimensions and fusing the obtained feature maps per dimension. We show that our model is able to predict multiple steps ahead with smaller error in comparison to the other examined models.

This paper is organised as follows. A brief overview of related research works for weather forecasting is given in section \ref{sec:related_work}. The proposed deep CNN-based architecture for wind speed forecasting together with the models that we used for comparison are presented in section \ref{sec:methods}. Section \ref{sec:experiments} describes the conducted experiments and the obtained results. A discussion on models performance is given in section \ref{sec:discussion}. We end with some conclusive remarks in section \ref{sec:conclusion}.
\section{Related Work}
\label{sec:related_work}
A number of different machine learning models have been proposed for weather forecasting. One of the most widely adopted approaches has seen Multilayer Perceptron (MLP) as main model for this purpose \cite{cadenas2009short, monfared2009new, li2010comparing, rohrig2006application}. An MLP is a rather simple neural network that consists of at least three layers: an input layer, a hidden layer and an output layer. Although MLPs were shown to be useful for a large range of applications, recently more sophisticated models such as convolutional neural networks have been applied to more challenging and complex tasks. Initially, CNNs have been used for image applications due to their aptness at capturing spatial relations \cite{krizhevsky2012imagenet}. However, since then, CNNs have also been successfully applied to time-series tasks. For instance, the authors in \cite{liang2019deep} utilised a CNN-based model to achieve high classification accuracy for a time-series task for detecting human transportation mode. The input to their model consists of 1-D data because they only have access to one sensor reading. Weather data, in contrast, often has multiple sensor readings resulting in multi-dimensional data, thus presenting a more complex time-series task that requires a more sophisticated model.

For weather prediction tasks, convolutions are often used in conjunction with long short-term memory (LSTM). An LSTM is a particular type of recurrent neural network which, unlike standard feedforward neural networks, includes feedback connections which provide the network with some kind of memory of previous signals, making it adept at learning from experience and well-suited for predictions using time series data \cite{hochreiter1997long}. Taking advantage of this, the authors in their seminal work in \cite{xingjian2015convolutional} were the first to combine convolutions with an LSTM to create the ConvLSTM model for precipitation forecasting. An important challenge in machine learning-based precipitation forecasting is that the rain maps that are typically used as input are high-dimensional data which requires the model to make efficient use of the available information. Extending on the idea of \cite{xingjian2015convolutional}, the authors in \cite{wang2017predrnn} therefore stacked multiple convolutional LSTMs while also adding extra memory connections in order to enable efficient flow of spatial information, resulting in their PredRNN model.

In the field of precipitation forecasting, the authors in \cite{agrawal2019machine} employed a U-net architecture \cite{ronneberger2015u} to predict rainfall one hour ahead using the last hour of rain maps. A U-net is a type of CNN that employs an encoder-decoder structure which at first applies several iterations of downsampling followed by several iterations of upsampling. In addition, the upsampling integrates the output from previous layers through so-called skip-connections. This particular architecture combining up- and downsampling enables a synergistic learning of both low- and high-level features. An interesting aspect of \cite{agrawal2019machine} is that they predict four discrete classes of rainfall based on millimetres of rain per hour, rather than predicting exact values of rainfall as has predominantly been done in previous models. This classification approach has recently been extended by \cite{sonderby2020metnet} who map their output onto 512 bins, thereby resulting in a much finer resolution. 


A frequently used baseline in data-driven weather forecasting is the `Na\"{i}ve Predictor' which predicts for time $t+\Delta t$ the same value as at time $t$ \cite{soman2010review}. This method is also called `Persistence' because it is based on the assumption that the weather persists from one time window to the next. Although this prediction method is rather simple, it proved to be very accurate for very-short and short forecasts as described by \cite{soman2010review}. In the next section we describe the models that we use and explain our proposed model architecture that was applied to real-life data for the wind speed prediction task. 

\section{Methods}
\label{sec:methods}
\subsection{Proposed multidimensional model}
The present study builds upon and extends the work of \cite{mehrkanoon2019deep} who introduced different CNN architectures, including 1-D, 2-D and 3-D convolutions, to accurately predict wind speed in the next 6 to 12 hours. In particular, for our new model, we apply depthwise-separable convolutions (DSCs) \cite{chollet2017xception} to all three input tensor dimensions (channel, width, height). 
In this way, the kernels of the three convolutions go over the height-and-width, channel-and-height and channel-and-width dimensions, respectively. For the studied wind speed prediction task, this corresponds to time steps-features, cities-features, cities-time steps. The output of the three DSCs are then concatenated and fed into a fully connected layer. Thanks to the proposed architecture, the underlying information within each view of the input are learnt jointly in an end-to-end fashion. 

\begin{figure*}
	\centering
		\includegraphics[width=.95\textwidth]{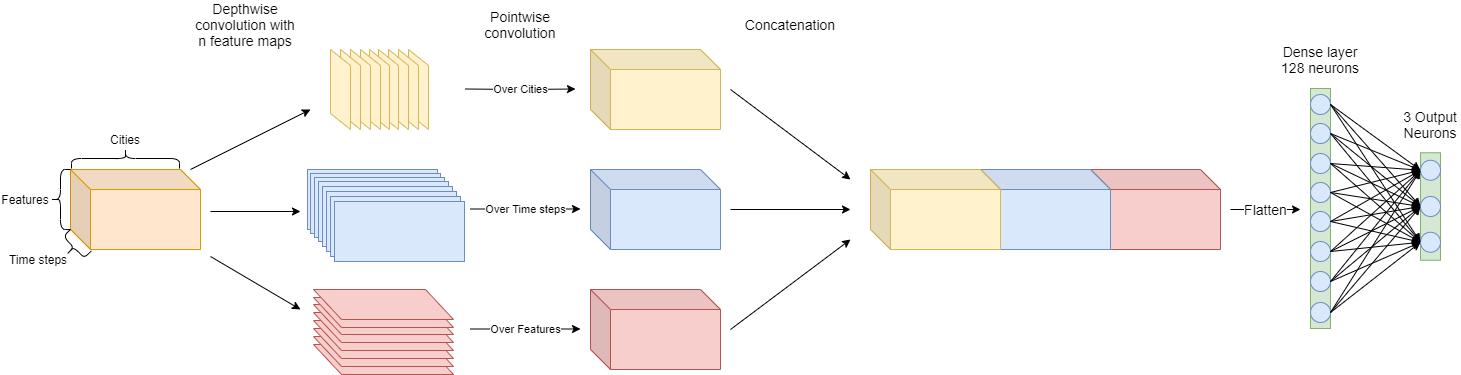}
	\caption{Using depthwise-separable convolutions to go over different dimensions of the input tensor. The resulting three tensors are concatenated along the city-dimension and then flattened and fed into a dense layer with 128 hidden neurons. Lastly, a dense layer with three neurons calculates the final output.}
	\label{fig:tensor}
\end{figure*}

Depthwise-separable convolutions have their convolution operation split up into two convolutions: a depthwise and a pointwise convolution. The advantage of using DSCs over normal convolutions is that they require a significant amount less parameters while keeping a similar accuracy \cite{lawhern2018eegnet}. Here, we train our model to predict the wind speed multiple steps ahead. As the prediction time ahead increases, the prediction task becomes increasingly complex for the model. After applying DSCs to all three dimensions we flatten the resulting feature maps to feed it into a fully connected layer followed by the output layer. In order to be able to apply a convolution to other dimensions, the input tensor needs to be permuted. A visual interpretation of this approach can be found in Fig.~\ref{fig:tensor}. In our approach we set the number of feature maps per channel to 16 and the number of hidden neurons of the fully connected layer to 128. Following every pointwise convolution we apply a batch normalisation. A rectified linear (ReLu) activation function \cite{nair2010rectified} is used after batch normalisation as well as after the dense layer. 

\subsection{Alternative convolutional models}
We have compared the performance of our proposed model with four alternative CNN-based models.  
This includes the 2D and 3D convolutional models from \cite{mehrkanoon2019deep}. A 2D convolutional approach using attention \cite{bello2019attention} and a 2D convolution with a transposed convolution of the original input followed by a double convolution as it is used in a U-net structure \cite{ronneberger2015u}, but instead of normal convolutions we used depthwise-separable convolutions. A transposed convolution is used for upscaling the input dimensions. It has a similar effect than upsampling, but with a transposed convolution the network can adjust the kernels for the upsampling as well, allowing it to better adapt to the underlying data. The models with their respective trainable parameters can be found in Table \ref{tab:models}. The amount of parameters differed for the two datasets, because their input dimensions were of different size. 

\begin{figure*}
    \centering
    \subfloat[]{\includegraphics[width=.95\textwidth]{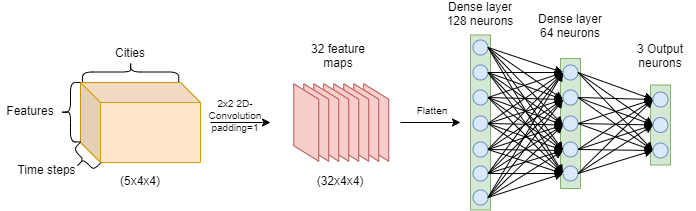}}
    
    \subfloat[]{\includegraphics[width=.95\textwidth]{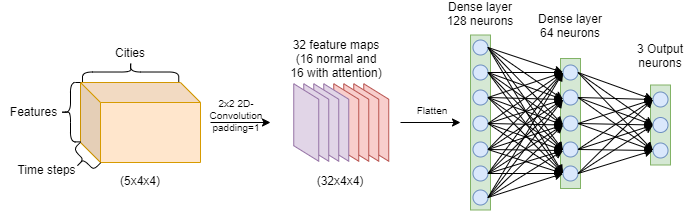}}
    
    \subfloat[]{\includegraphics[width=.95\textwidth]{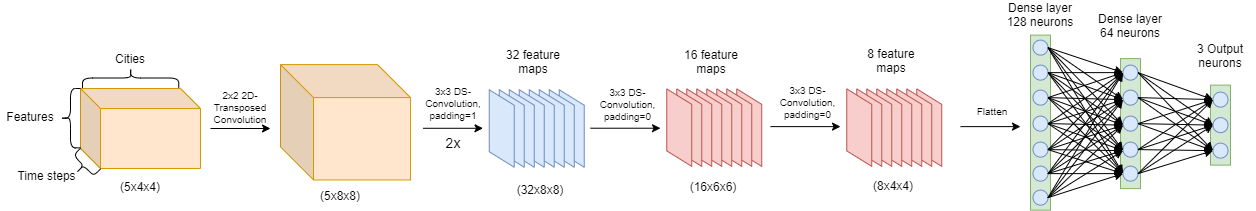}}
    
    \subfloat[]{\includegraphics[width=.95\textwidth]{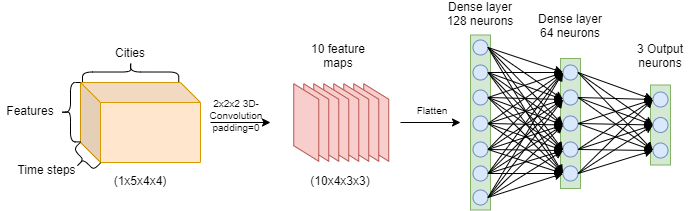}}
    \caption{The model architecture of the models that were tested. (a) 2D convolutional neural network. (b) 2D convolutional neural network with attention. (c) 2D convolutional neural network with transposed convolution (upscaling). (d) 3D convolutional neural network. The input dimensions and number of output neurons are adjusted to the dataset. Here, the models for the DK dataset are shown.}
    \label{fig:models}
\end{figure*}

\begin{table}[h]
    \renewcommand{\arraystretch}{1.1}
    \caption{The number of parameters of the five models that were compared in the experiments.}
    \label{tab:models}
    \centering
    \begin{tabular}{|c|c|c|c|}
    \hline
         \textbf{Model} & \textbf{\# Feature maps} & \multicolumn{2}{|c|}{\textbf{Parameters}} \\
         \hline
         & & \textit{Denmark} & \textit{Netherlands} \\
         \hline
         2D & 32 & 46,115 & 112,167 \\
         2D + Attention & 32 & 47,059 & 113,367\\
         2D + Upscaling & 32 & 27,974 & 77,568 \\
         3D & 10 & 54,749 & 200,929 \\
         Multidimensional & 16 & 37,258 & 102,832 \\
         \hline
    \end{tabular}

\end{table}
It can be seen from Table \ref{tab:models} that there is a large difference between the number of parameters in the models of the Dutch and Danish dataset. This difference arises from different input sizes of the dataset. The Danish dataset has an input shape of $5 \times 4 \times 4$ and the Dutch dataset has an input shape of $7 \times 6 \times 6$ (see section \ref{sec:experiments} for a more detailed description of the datasets). Since we are flattening the output of the convolutional layers and feed it to a dense layer, the number of parameters increase dramatically. The architectures of the alternative models used for comparison are shown in Fig.~\ref{fig:models}.  

The number of feature maps for the 2D and 3D CNN-based models are set to 32 and 10, respectively. For our proposed model the number of feature maps is set to 16 for every dimension.
We train the models for a maximum of 150 epochs with a batch size of 64. An early stopping is used to track the validation error and if it does not decrease after 20 epochs the training will stop. We use the Adam optimiser \cite{kingma2014adam} with default values and a learning rate of 0.001 for updating the model parameters. 
The code of our proposed model and the other trained models along with the datasets can be found on Github\footnote{\url{https://github.com/HansBambel/multidim_conv}}.
\section{Experiments}
\label{sec:experiments}
\begin{figure*}[!t]
    \centering
    \subfloat[Danish weather stations]{\includegraphics[width=.35\textwidth]{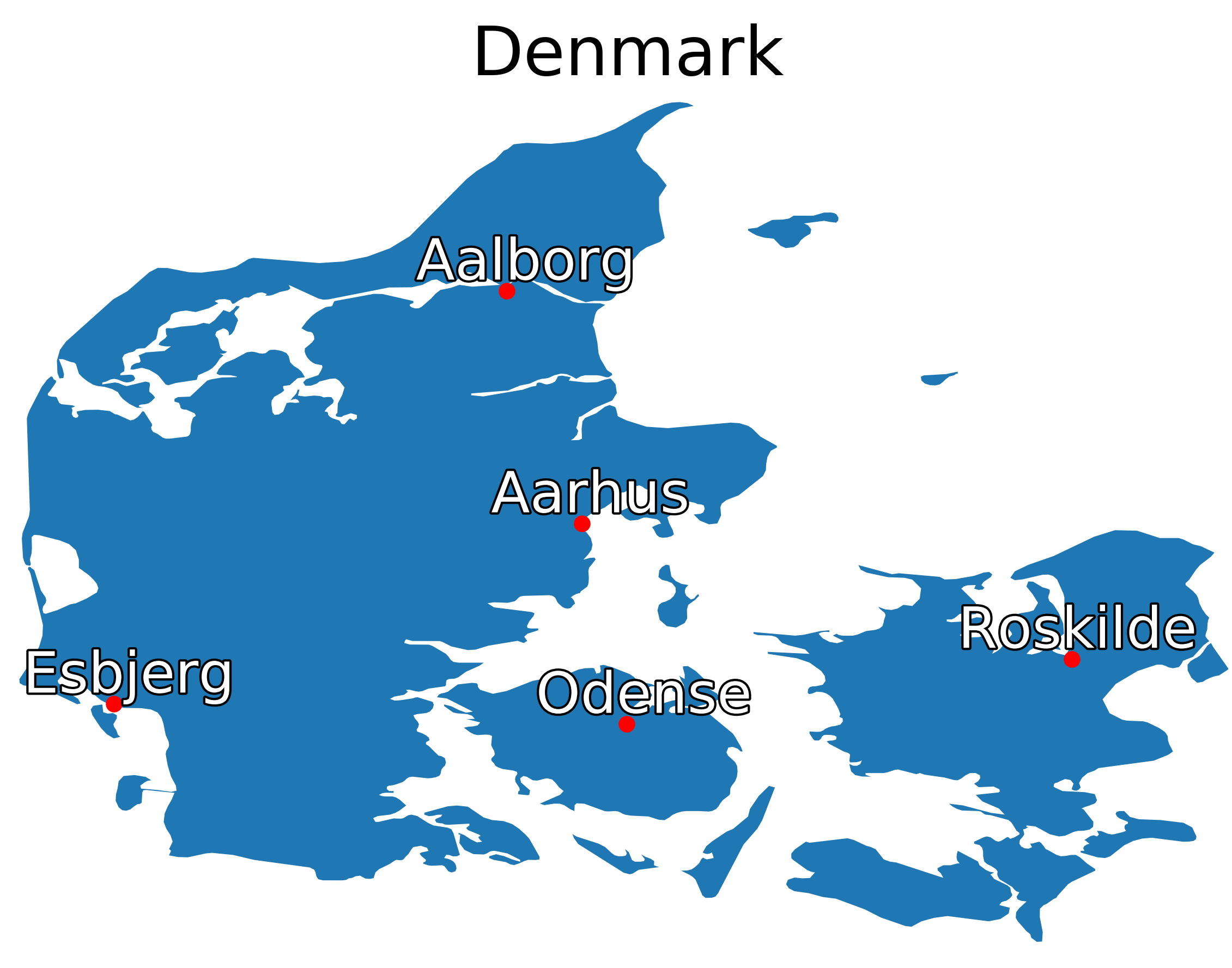}}
    \hspace{60px}
    \subfloat[Dutch weather stations]{\includegraphics[width=.35\textwidth]{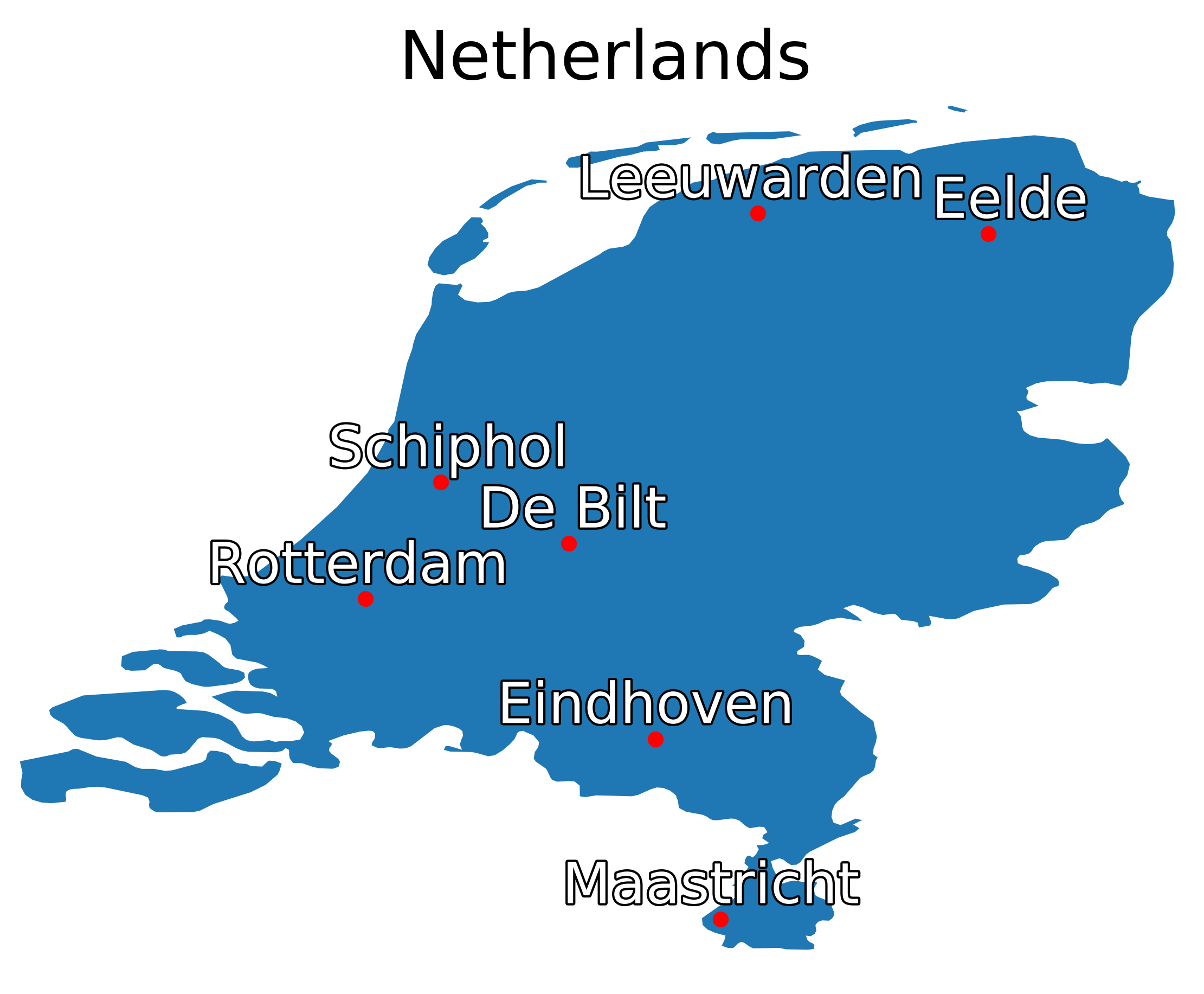}}
    \caption{The cities where the wind data was taken: (a) Aalborg, Aarhus, Esbjerg, Odense and Roskilde. (b) Schiphol, De Bilt, Leeuwarden, Eelde, Rotterdam, Eindhoven, Maastricht.}
    \label{fig:cities}
\end{figure*}

We evaluate our model together with the other discussed four models on two different real-life datasets from Denmark and the Netherlands. The cities from which the weather measurements have been recorded and used in our experiments are shown in Fig.~\ref{fig:cities}. 
The performance of the models is measured by their mean absolute error (MAE) and mean squared error (MSE) on the test set. MAE calculates the mean of the difference between the predicted values and true values and is calculated as follows:
\begin{equation}
    MAE = \frac{\sum^n_{i=1} |y_i - \hat{y}_i|}{n},
\end{equation}
where $n$ is the number of samples and $y_i$ and $\hat{y}_i$ is the true and predicted value, respectively.
MSE calculates the mean of the squared difference between the predicted values and the true values. As a result, larger differences between $y$ and $\hat{y}$ contribute more to the error term than smaller differences. MSE is calculated as follows:
\begin{equation}
    MSE = \frac{\sum^n_{i=1} (y_i - \hat{y}_i)^2}{n}.
\end{equation}
MAE was chosen to be able to quantitatively compare the results with those in \cite{mehrkanoon2019deep} and MSE was chosen to have an indicator how reasonable the predictions of the models are. In what follows, the two used datasets are described in more detail.

\subsection{Dataset Denmark}
This is the same dataset as used in \cite{mehrkanoon2019deep} and available online\footnote{\url{https://sites.google.com/view/siamak-mehrkanoon/code-data}}. It contains hourly measurements of temperature, pressure, wind speed and wind direction of five cities in Denmark (see Fig.~\ref{fig:cities}a) from 2000-2010. The dataset is split into training and validation (years 2000-2009) and test set (year 2010). The dataset is normalised using a min-max scaler that was trained on the training set. Furthermore, the dataset is arranged in a way that a sample contains the last four measurements in all the five cities and the target is the wind speed in three of the cities (Esbjerg, Odense and Roskilde). Therefore, the input dimensions have the shape $C \times T\times F$ where $C = \#cities$, $T = \#timesteps$ and $F = \#features$ (resulting in a shape of size $5 \times 4 \times 4$). Those are the three input dimensions of the left-most cube in Fig.~\ref{fig:tensor}. We trained all five models multiple times to predict the wind speed  multiple steps ahead which corresponds to 6, 12, 18 and 24 hours ahead in the Danish dataset. Additionally, we created a persistence model as baseline that uses the last wind speed measurement of each city as the prediction.

The average of the obtained MAEs and MSEs, corresponding to different models, over three target cities in Denmark are tabulated in Table \ref{tab:errors_DK}. The best MAE and MSE are underlined for each tested prediction time in Table \ref{tab:errors_DK}. Our model performs best or second best for all of the tested prediction time. In particular, one can observe that our proposed multidimensional architecture has the smallest overall MAE in the 6h and 24h prediction and the smallest MSE in the 6h and 24h prediction. The error increases in all models the further ahead into the future the model needs to predict. This makes sense because values in the near future are less likely to change than those further away, therefore they are easier to predict. 
\begin{table*}
    \centering
    \caption{The average MAEs and MSEs of different models over target cities in Denmark dataset.  
    }
    \label{tab:errors_DK}
    \begin{tabular}{|c|c|c|c|c|c|c|c|c|}
    \hline
        \textbf{Model} & \multicolumn{4}{|c|}{\textbf{MAE}} & \multicolumn{4}{|c|}{\textbf{MSE}}\\
        \hline
          & \textit{6h} & \textit{12h} & \textit{18h} & \textit{24h} & \textit{6h} & \textit{12h} & \textit{18h} & \textit{24h}\\
         \hline
        Persistence     &  1.649  & 2.210  & 2.309  &  2.313 & 4.608  & 7.929  & 8.702  & 8.812 \\
        2D              &  1.304  & 1.746  & 1.930  &  2.004 & 2.824  & 5.088  & 6.120  & 6.610 \\
        2D+Attention    &  1.313  & 1.715  & 1.905  &  1.950 & 2.885  & 4.896  & 5.933  & 6.201 \\
        2D+Upscaling    &  1.307  & 1.723  & \underline{1.858}  &  1.985 & 2.826  & 4.931  & \underline{5.639}  & 6.474 \\
        3D              &  1.311  & \underline{1.677}  & 1.908  &  1.957 & 2.855  & \underline{4.595}  & 5.958  & 6.238 \\
        Multidimensional&  \underline{1.302}  & 1.706  & 1.873  &  \underline{1.925} & \underline{2.804}  & 4.779  & 5.773  & \underline{6.066}  \\
        \hline
    \end{tabular}
    
\end{table*}
Furthermore, one can note that the error increases less in the proposed multidimensional model than in the other models. This is an indicator that our model has better generalisation capabilities and can adapt better to different kinds of data. 
The baseline model is outperformed by a large margin in all step ahead predictions.

\begin{figure*}
    \centering
    \subfloat[6 hours ahead]{
        \includegraphics[width=.4\textwidth]{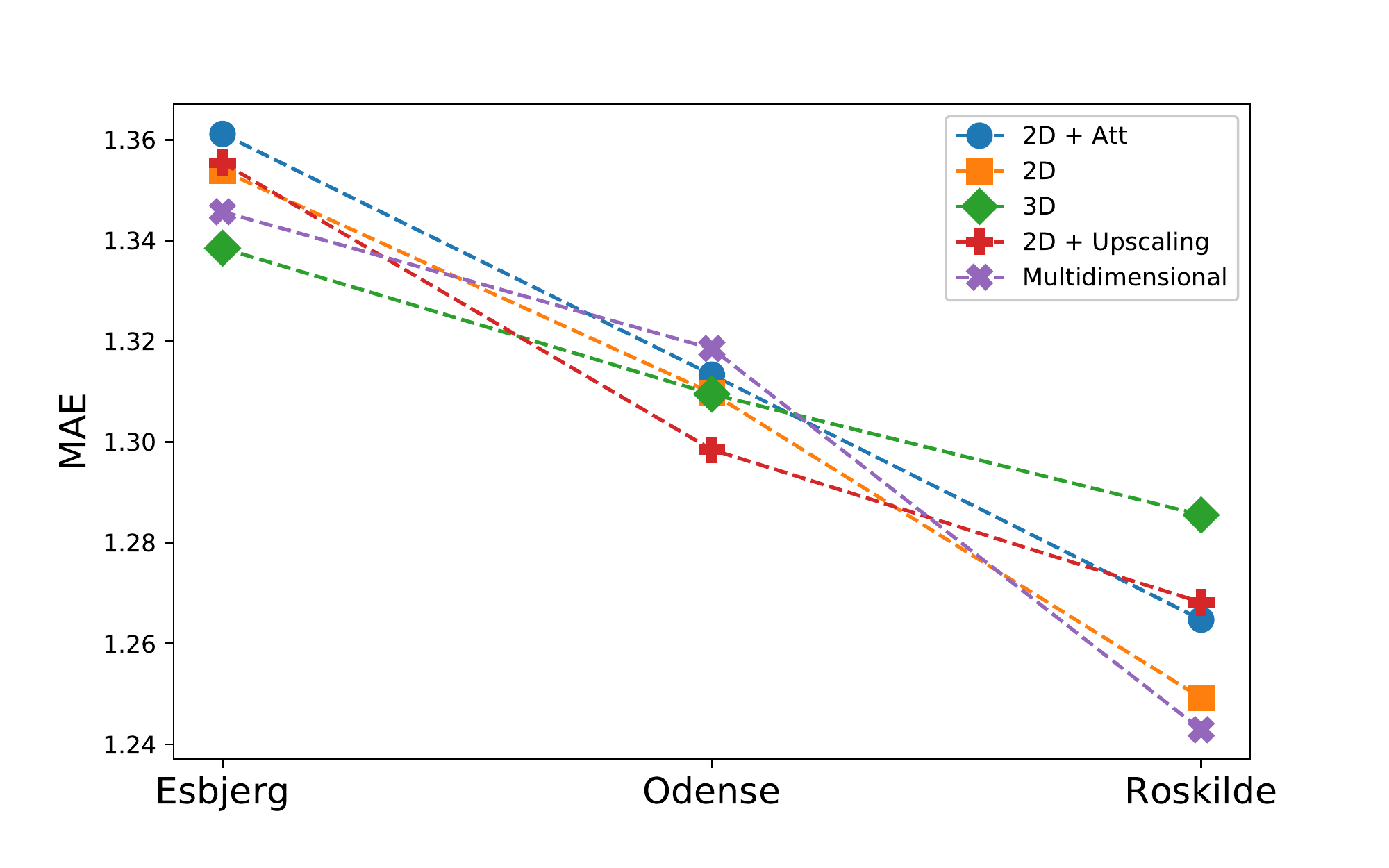}}\quad
    \subfloat[12 hours ahead]{
        \includegraphics[width=.4\textwidth]{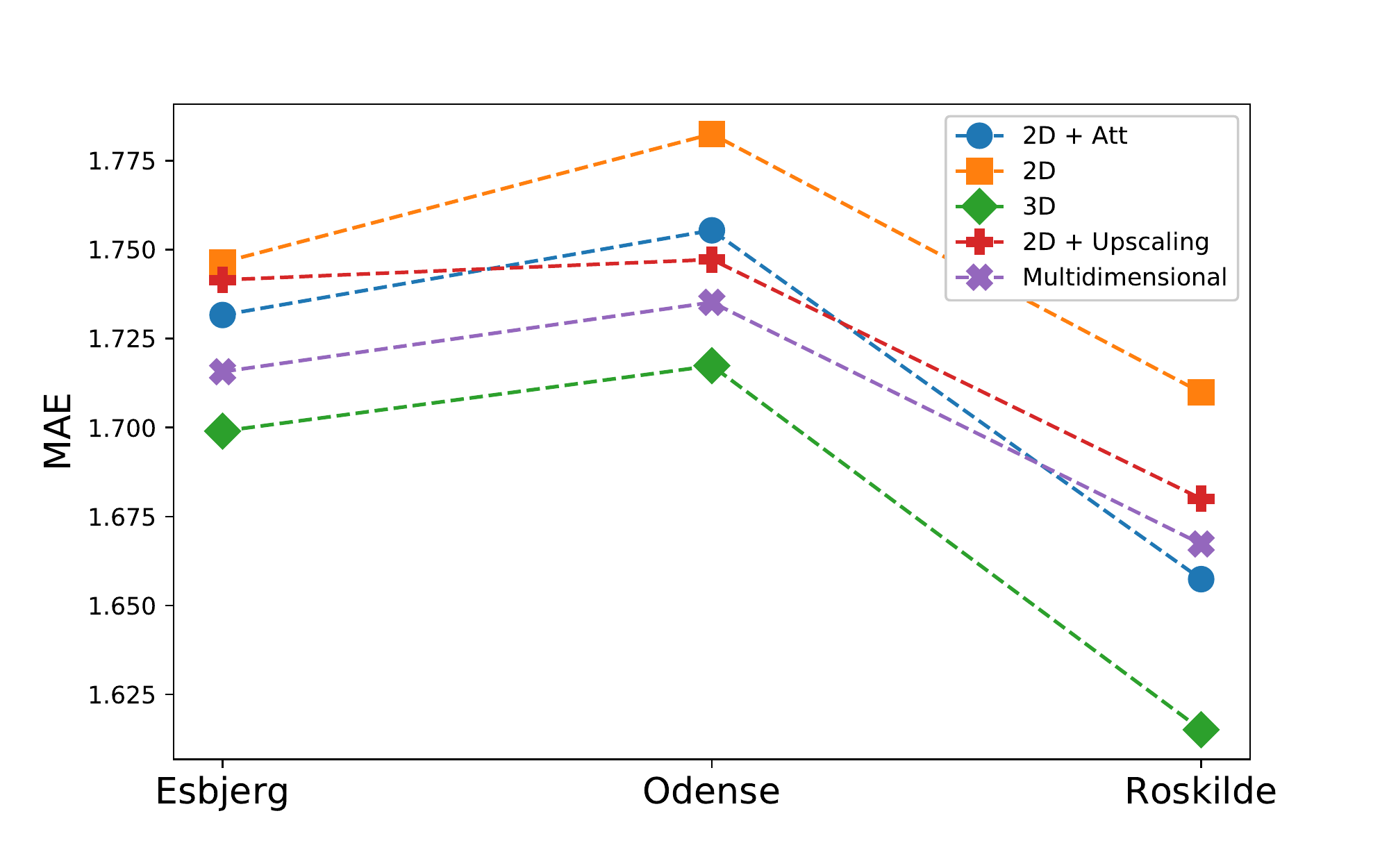}}\\
    \subfloat[18 hours ahead]{
        \includegraphics[width=.4\textwidth]{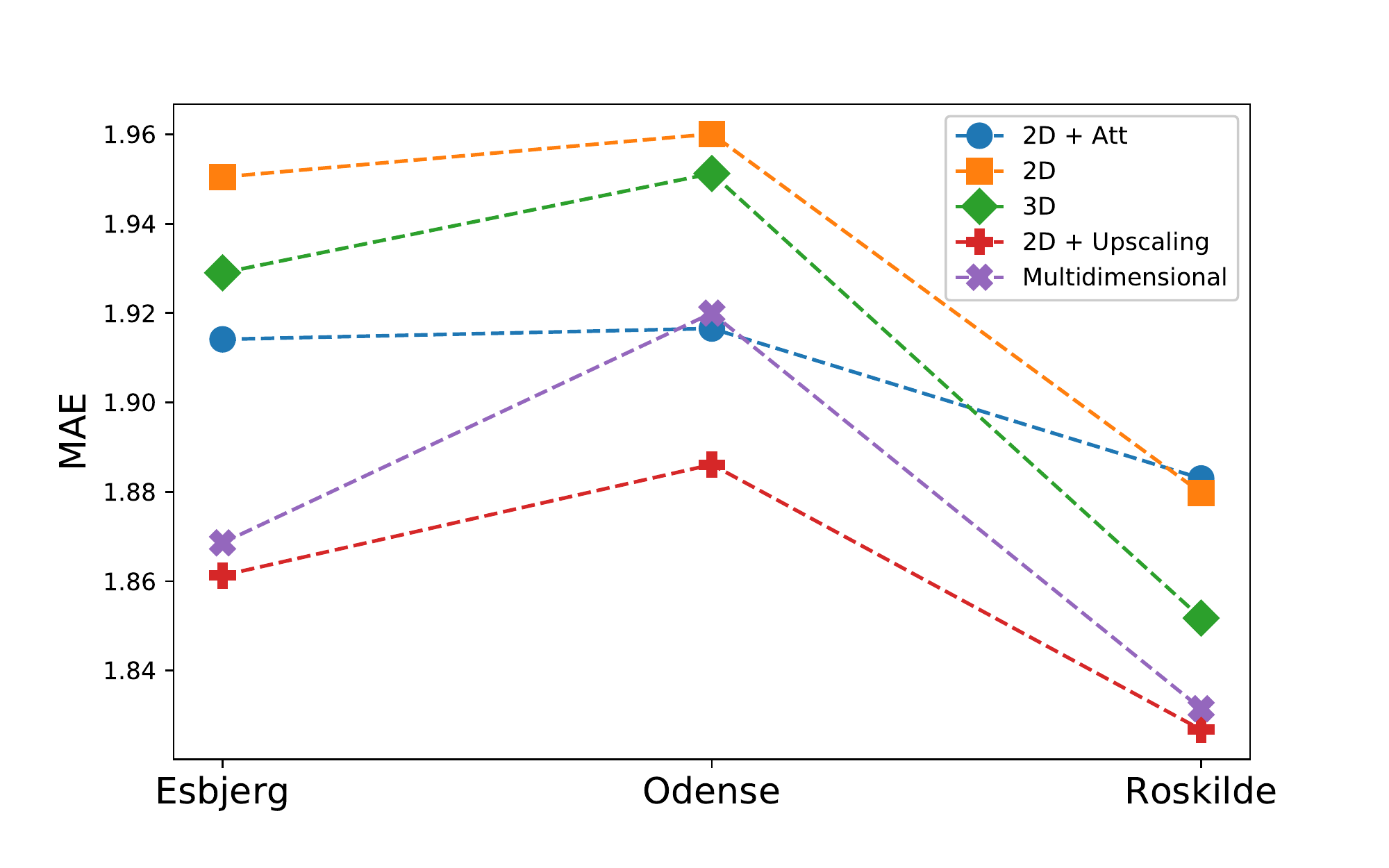}}\quad
    \subfloat[24 hours ahead]{
        \includegraphics[width=.4\textwidth]{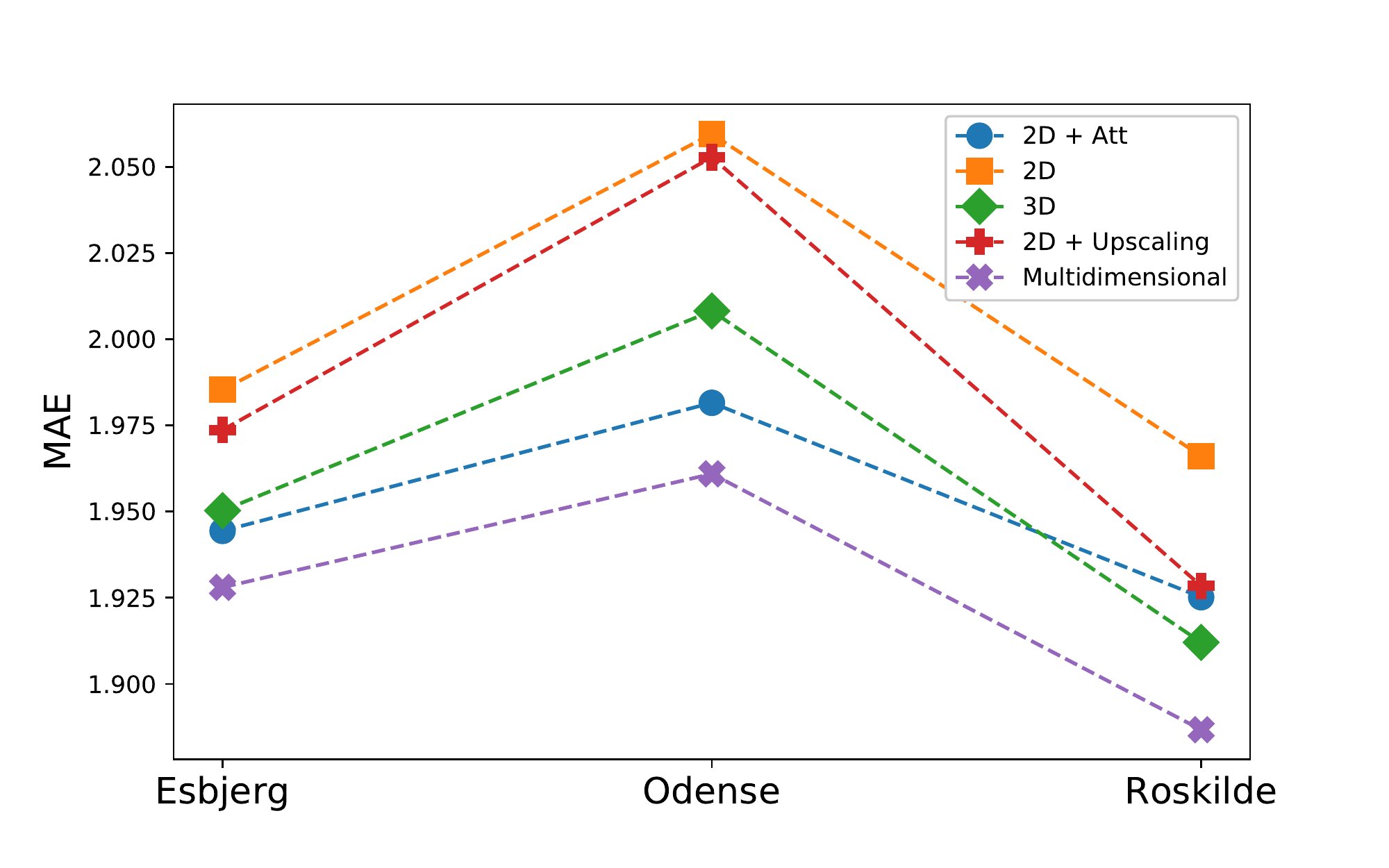}}
    \caption{The models performance for different time steps ahead for each cities forecast of the Denmark dataset.} 
    \label{fig:pred_cities}
\end{figure*}

\begin{figure}
    \centering
    \includegraphics[width=.4\textwidth]{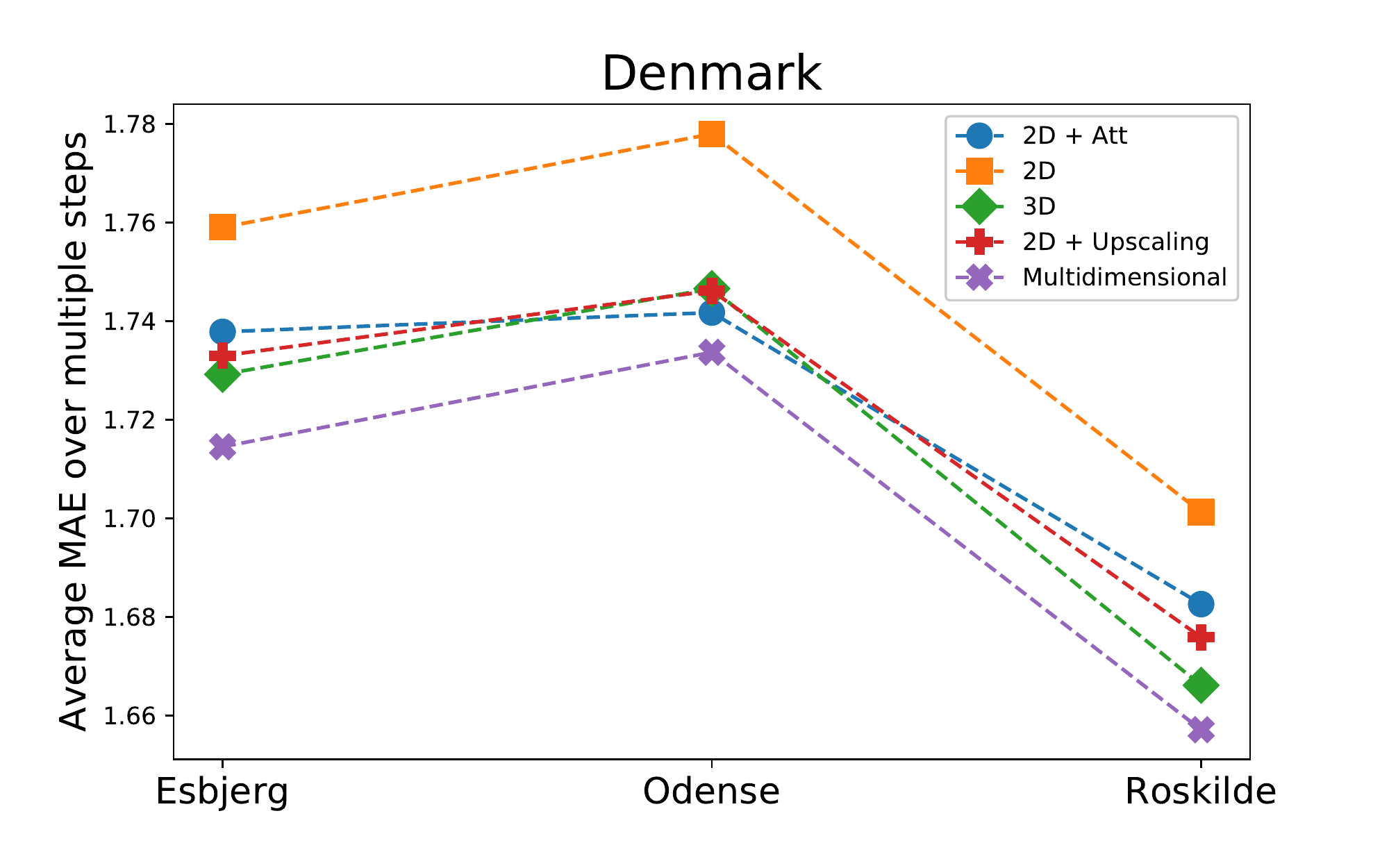}
    \caption{Mean of the models performance across different time steps for each city in the Danish dataset.}
    \label{fig:error_cities_dk}
\end{figure}

In Fig.~\ref{fig:pred_cities} the MAE of the models for each of the three target cities in all four prediction times are shown. The plots show that our model has, for almost all time steps, the lowest or second lowest error in each city. The differences in the performance of the models get larger the further they have to predict into the future. 
To illustrate further how well the models are performing across all time steps we calculated the mean of the MAEs for each city across the time steps (see Fig.~\ref{fig:error_cities_dk}). We can observe that our model is performing best for each of the three cities.

\subsection{Dataset Netherlands}
The second dataset for this study was acquired from the Royal Netherlands Meteorological Institute. It contains hourly weather measurements from 7 cities in the Netherlands (Fig.~\ref{fig:cities}b) from January 1, 2011 until March 29, 2020 resulting in 81.000 data points. The measured weather features are wind speed, wind direction, temperature, dew point, air pressure and rain amount. We split the dataset into training (January 2011 - December 2018) and testing (January 2019 - March 2020). The arrangement of the dataset is the same as the Denmark dataset: $C \times T \times F$. We choose to use the last 6 measurements as input and predict the wind speed 1-, 2-, 3- and 4-steps ahead (corresponding to 1h, 2h, 3h and 4h ahead, respectively), which results in an input tensor of shape $7 \times 6 \times 6$. Additionally, we create a persistence baseline by predicting the last given input values. 

The performance of the models trained on this dataset can be found in Table \ref{tab:errors_NL}. From Table \ref{tab:errors_NL}, one can observe that our model has the best performance in the 2h and 3h prediction when looking at the MAE and outperforms the other models in the 2h, 3h and 4h ahead predictions when using MSE. The baseline is outperformed by every model by a big margin. 

\begin{table*}
    \centering
    \caption{The average MAEs and MSEs of different models over target cities in the Netherlands dataset.}
    \label{tab:errors_NL}
    \begin{tabular}{|c|c|c|c|c|c|c|c|c|}
    \hline
        \textbf{Model} & \multicolumn{4}{|c|}{\textbf{MAE}} & \multicolumn{4}{|c|}{\textbf{MSE}}\\
        \hline
          & \textit{1h} & \textit{2h} & \textit{3h} & \textit{4h} & \textit{1h} & \textit{2h} & \textit{3h} & \textit{4h} \\
         \hline
        Persistence  &  9.55 & 11.34 & 12.90 & 14.37  & 183.61 & 246.95 & 310.38 & 375.36 \\
        2D           &  8.11 & 9.17  & 10.15 & 11.12 & 116.89 & 149.01 & 181.78 & 218.49 \\
        2D+Attention &  \underline{8.08} & 9.10  & 10.11 & 11.00  & \underline{115.96} & 147.75 & 180.66 & 213.23 \\
        2D+Upscaling &  8.16 & 9.07  & 10.14 & \underline{10.85}  & 117.80 & 147.21 & 182.44 & 208.96\\
        3D           &  8.17 & 9.26  & 10.15 & 10.93   & 118.35 & 151.51 & 181.35 & 211.19\\
        Multidimensional     &  8.12 & \underline{9.05}  & \underline{9.95}  & 10.94   & 116.78 & \underline{144.51} & \underline{174.07} & \underline{208.73} \\
        \hline
    \end{tabular}
\end{table*}

\begin{figure}
    \centering
    \includegraphics[width=0.4\textwidth]{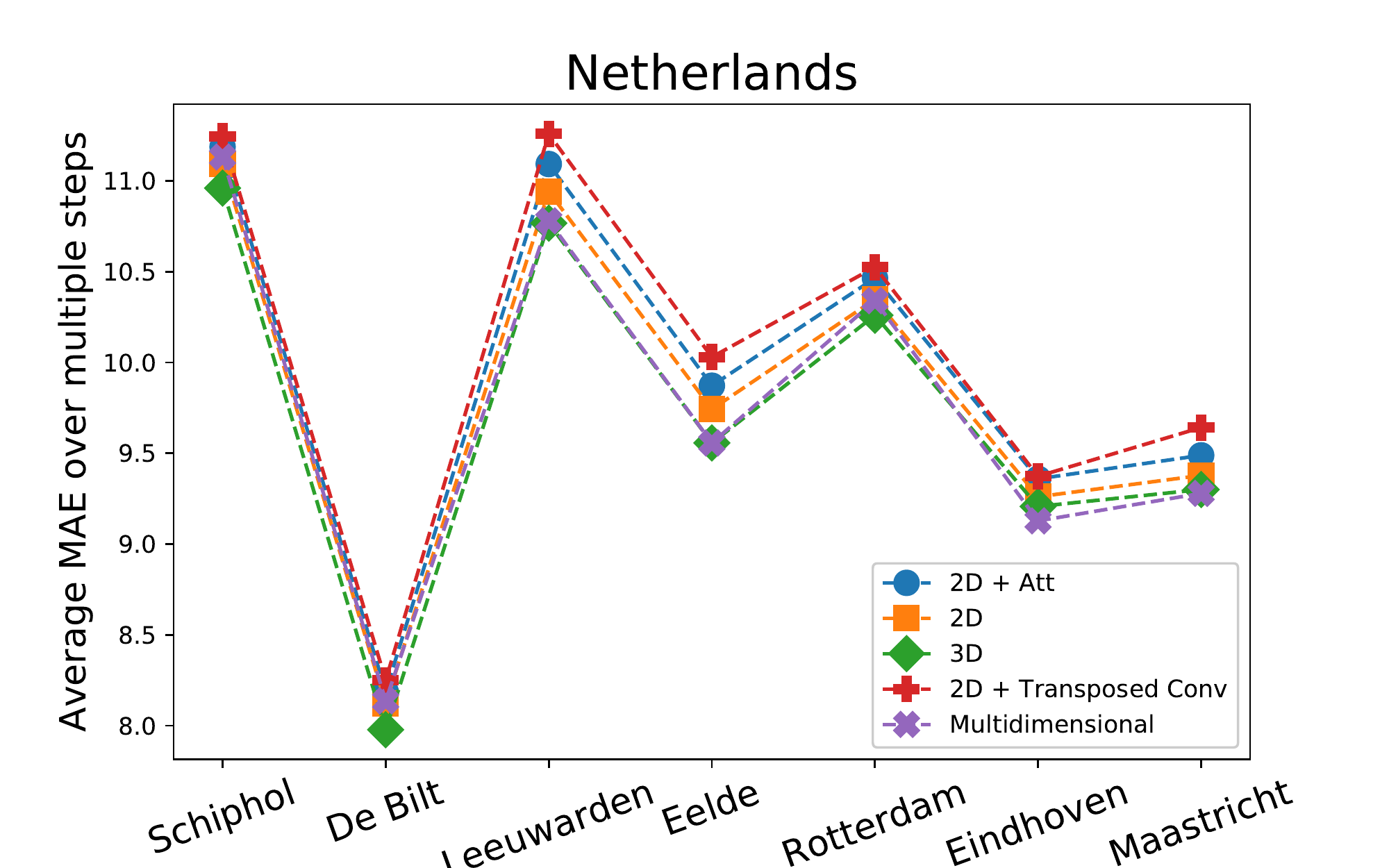}
    \caption{Mean of the models performance across different time steps for each city in the Dutch dataset.}
    \label{fig:error_cities_nl}
\end{figure}

\begin{figure*}
    \centering
    \subfloat[]{
        \includegraphics[width=.45\textwidth]{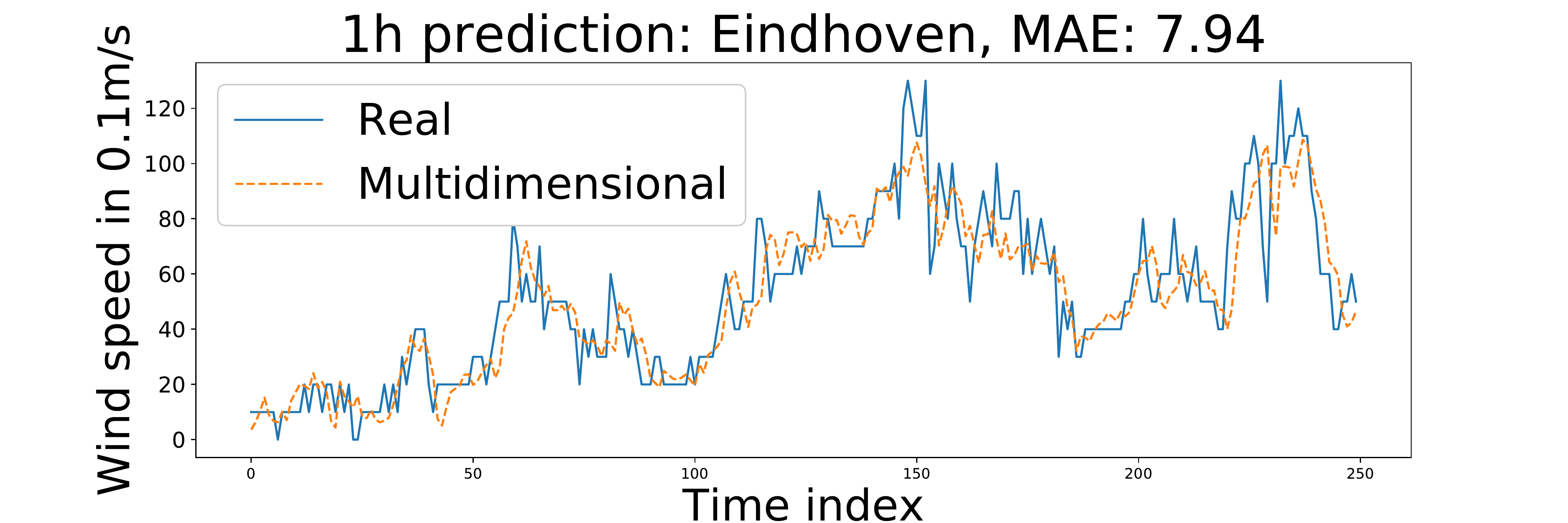}}\quad
    \subfloat[]{
        \includegraphics[width=.45\textwidth]{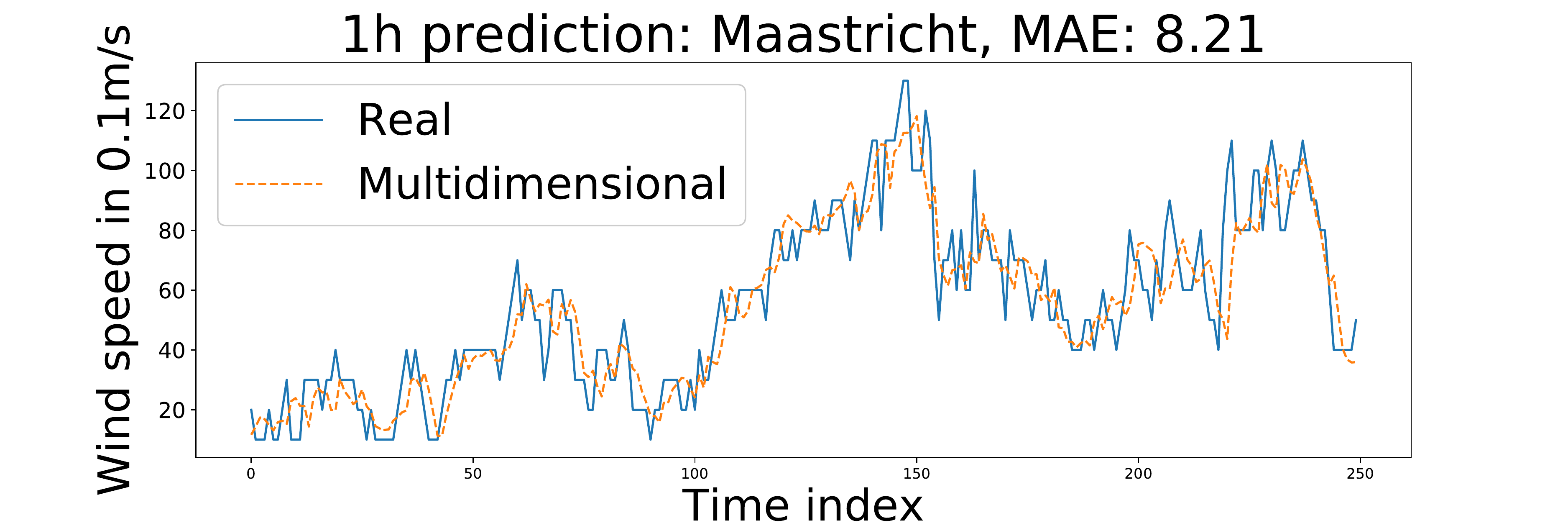}}\\
    \subfloat[]{
        \includegraphics[width=.45\textwidth]{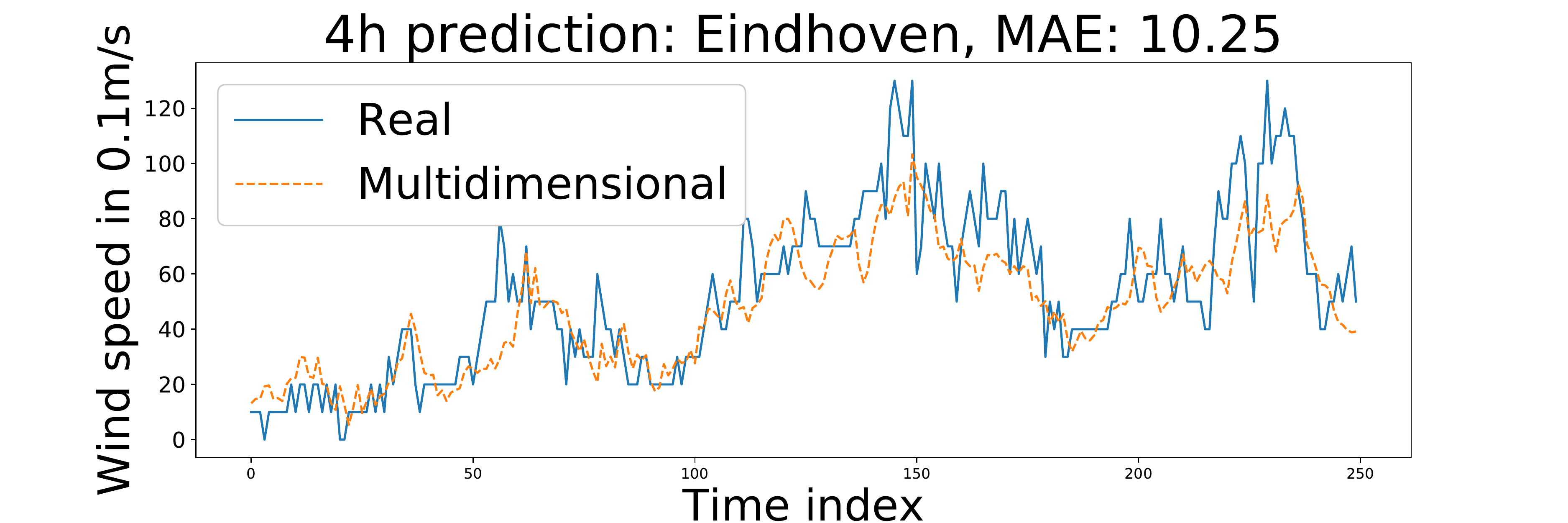}}\quad
    \subfloat[]{
        \includegraphics[width=.45\textwidth]{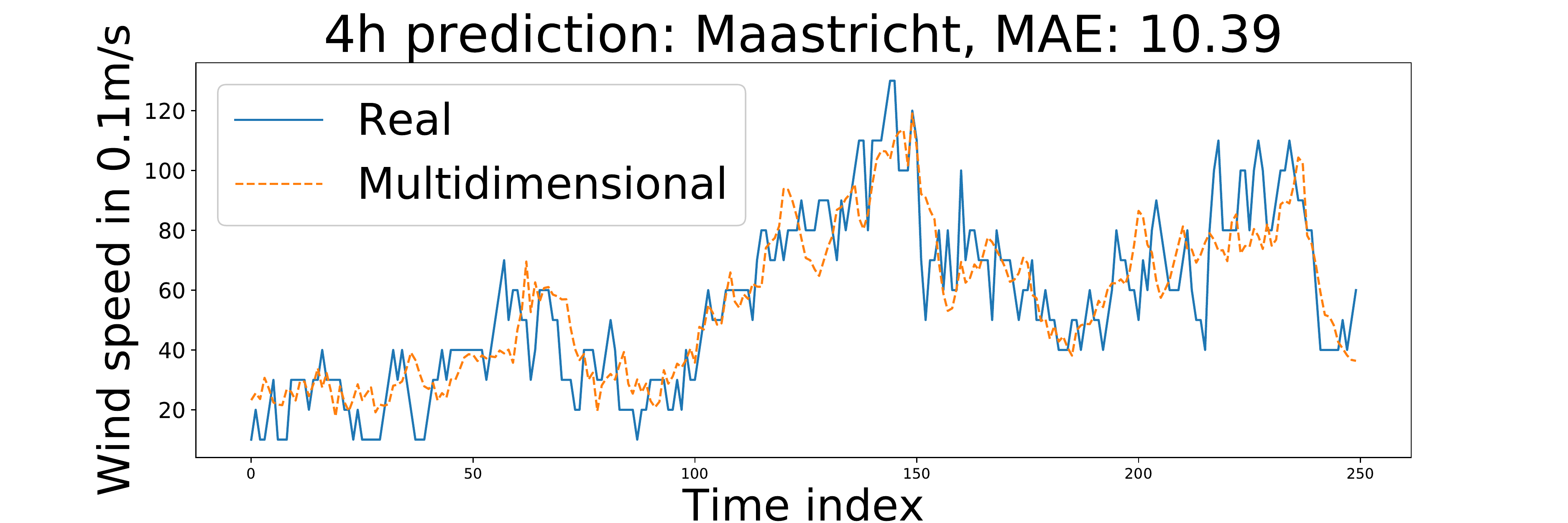}}
    \caption{Comparison of predictions using our model and the real value. It is visible that the 1 hour predictions (a) and (b) follow the trend more closely than the 4 hour predictions (c) and (d).} 
    \label{fig:example_pred}
\end{figure*}

For this dataset we also calculated the mean of the MAEs for each individual city across the four different time steps. The results are shown in Fig.~\ref{fig:error_cities_nl}. It can be seen that our multidimensional model performs best or second best in five out of the seven cities.
An example of a 1 hour and 4 hour prediction from our model is shown in Fig.~\ref{fig:example_pred}. For this, we used a subset of the data and plotted the wind speed prediction for the cities of Eindhoven and Maastricht at each time step. Note that the MAE reported is for the whole test set.
The Fig.~\ref{fig:example_pred} shows that 1-hour ahead prediction of our model more closely follows the real measurements compared to that of the 4-hours ahead predictions.
\section{Discussion}
\label{sec:discussion}
In this study, we propose a new convolution neural network for short-term wind speed predictions using weather datasets from Denmark and the Netherlands. Specifically, our new model applies depthwise-separable convolutions to three dimensions of the input tensor. To assess the model performance, we compare it to four previously described CNNs, including a 2D model, a 3D model, a 2D model with attention and a 2D model with transposed convolution.

Using MAE and MSE as performance indicators, the analysis of our experiments shows that our model outperforms previous prediction approaches for most time steps on both of the tested datasets. We hypothesise that these performance improvements can be attributed to our novel approach of applying convolutions separately over different input dimensions, rather than applying them to only the height and width dimensions as done by standard convolutional approaches. The proposed architecture enables the network to more efficiently exploit information from all dimensions.

Our findings of the 2D and 3D models are consistent with the results of \cite{mehrkanoon2019deep}. They showed that a 3D-convolutional approach outperforms a 2D-convolutional approach on the Danish dataset. Likewise, in our study, the 3D-convolution model performed better than the regular 2D-convolution model for all time steps on the Danish dataset except for the 6-hour prediction and thus, overall, performed better than the 2D-convolution model (see Table \ref{tab:errors_DK}). Interestingly, the 2D-convolutional models with attention and upscaling are able to achieve better results than the regular 2D- and 3D-models.

We did not find the same high performance of the 3D-convolutional model for the Dutch dataset when looking at different time steps. Indeed, when disregarding the baseline model, the 3D model showed the worst or second worst MAE and MSE for most time steps as can be seen in Table \ref{tab:errors_NL}. Interestingly, the 3D model performs well when comparing the mean of the time steps for each individual city (see Fig.~\ref{fig:error_cities_dk} and \ref{fig:error_cities_nl}). Our multidimensional model performed better than the other models in the later time step prediction when considering MSE as performance indicator. As previously mentioned, a smaller MSE is an indicator of how reasonable the model predicts. 
Overall, the high performance of our model generalises better across time steps and the two used datasets (see Fig.~\ref{fig:error_cities_dk} and \ref{fig:error_cities_nl}), largely providing better predictions than previously described standard based convolutional approaches.

An important limitation of our study concerns the comparability of the five models. We used number of parameters as main indicator for model comparability, rather than optimising every model individually which was beyond the scope of the present study. Thus, we sought to keep the number of parameters approximately similar in all models without modifying too many other configurations. To this end, we set the number of feature maps for the three 2D models to 32, for the 3D model to 10, and to 16 per dimension in our model (see Table \ref{tab:models}). The number of feature maps for the different models was based on the input dimensions of the Danish dataset and were subsequently also applied to the Dutch dataset that has larger input dimensions than the Danish dataset. This explains the greater range in number of parameters of the models when applied to the Dutch dataset (77,568 - 200,929 parameters) compared to the Danish dataset (27,974 - 54,749 parameters), as shown in Table \ref{tab:models}. We also trained the 2D models with less feature maps (10 feature maps, as in \cite{mehrkanoon2019deep}), but observed no performance increase.

\section{Conclusion and Future Work}
\label{sec:conclusion}
In this paper a new multidimensional CNN based model is introduced for wind speed prediction. 
We showed that applying convolutions to multiple dimensions helps to capture more modalities of the data. Using real weather datasets we tested our approach against conventional 2D- and 3D-CNNs and showed that our proposed model is able to more accurately predict wind speeds multiple steps ahead. Furthermore, our proposed multidimensional model performed best across different time steps compared to the other tested models. 

\bibliographystyle{IEEEtran}
\bibliography{IEEEabrv.bib,main}

\begin{thebibliography}{10}
\providecommand{\url}[1]{#1}
\csname url@samestyle\endcsname
\providecommand{\newblock}{\relax}
\providecommand{\bibinfo}[2]{#2}
\providecommand{\BIBentrySTDinterwordspacing}{\spaceskip=0pt\relax}
\providecommand{\BIBentryALTinterwordstretchfactor}{4}
\providecommand{\BIBentryALTinterwordspacing}{\spaceskip=\fontdimen2\font plus
\BIBentryALTinterwordstretchfactor\fontdimen3\font minus
  \fontdimen4\font\relax}
\providecommand{\BIBforeignlanguage}[2]{{%
\expandafter\ifx\csname l@#1\endcsname\relax
\typeout{** WARNING: IEEEtran.bst: No hyphenation pattern has been}%
\typeout{** loaded for the language `#1'. Using the pattern for}%
\typeout{** the default language instead.}%
\else
\language=\csname l@#1\endcsname
\fi
#2}}
\providecommand{\BIBdecl}{\relax}
\BIBdecl

\bibitem{li2010comparing}
G.~Li and J.~Shi, ``On comparing three artificial neural networks for wind
  speed forecasting,'' \emph{Applied Energy}, vol.~87, no.~7, pp. 2313--2320,
  2010.

\bibitem{soman2010review}
S.~S. Soman, H.~Zareipour, O.~Malik, and P.~Mandal, ``A review of wind power
  and wind speed forecasting methods with different time horizons,'' in
  \emph{North American Power Symposium 2010}.\hskip 1em plus 0.5em minus
  0.4em\relax IEEE, 2010, pp. 1--8.

\bibitem{agrawal2019machine}
S.~Agrawal, L.~Barrington, C.~Bromberg, J.~Burge, C.~Gazen, and J.~Hickey,
  ``Machine learning for precipitation nowcasting from radar images,''
  \emph{arXiv preprint arXiv:1912.12132}, 2019.

\bibitem{sonderby2020metnet}
C.~K. S{\o}nderby, L.~Espeholt, J.~Heek, M.~Dehghani, A.~Oliver, T.~Salimans,
  S.~Agrawal, J.~Hickey, and N.~Kalchbrenner, ``Metnet: A neural weather model
  for precipitation forecasting,'' \emph{arXiv preprint arXiv:2003.12140},
  2020.

\bibitem{mehrkanoon2014parameter}
S.~Mehrkanoon, S.~Mehrkanoon, and J.~A. Suykens, ``Parameter estimation of
  delay differential equations: an integration-free ls-svm approach,''
  \emph{Communications in Nonlinear Science and Numerical Simulation}, vol.~19,
  no.~4, pp. 830--841, 2014.

\bibitem{mehrkanoon2019cross}
S.~Mehrkanoon, ``Cross-domain neural-kernel networks,'' \emph{Pattern
  Recognition Letters}, vol. 125, pp. 474--480, 2019.

\bibitem{mehrkanoon2018deep}
S.~Mehrkanoon and J.~A. Suykens, ``Deep hybrid neural-kernel networks using
  random fourier features,'' \emph{Neurocomputing}, vol. 298, pp. 46--54, 2018.

\bibitem{mehrkanoon2017regularized}
------, ``Regularized semipaired kernel \uppercase{CCA} for domain
  adaptation,'' \emph{IEEE Transactions on Neural Networks and Learning
  Systems}, vol.~29, no.~7, pp. 3199--3213, 2017.

\bibitem{sapankevych2009time}
N.~I. Sapankevych and R.~Sankar, ``Time series prediction using support vector
  machines: a survey,'' \emph{IEEE Computational Intelligence Magazine},
  vol.~4, no.~2, pp. 24--38, 2009.

\bibitem{van2001financial}
T.~Van~Gestel, J.~A. Suykens, D.-E. Baestaens, A.~Lambrechts, G.~Lanckriet,
  B.~Vandaele, B.~De~Moor, and J.~Vandewalle, ``Financial time series
  prediction using least squares support vector machines within the evidence
  framework,'' \emph{IEEE Transactions on neural networks}, vol.~12, no.~4, pp.
  809--821, 2001.

\bibitem{scher2018toward}
S.~Scher, ``Toward data-driven weather and climate forecasting: Approximating a
  simple general circulation model with deep learning,'' \emph{Geophysical
  Research Letters}, vol.~45, no.~22, pp. 12--616, 2018.

\bibitem{hossain2015forecasting}
M.~Hossain, B.~Rekabdar, S.~J. Louis, and S.~Dascalu, ``Forecasting the weather
  of nevada: A deep learning approach,'' in \emph{2015 international joint
  conference on neural networks (IJCNN)}.\hskip 1em plus 0.5em minus
  0.4em\relax IEEE, 2015, pp. 1--6.

\bibitem{rodrigues2018deepdownscale}
E.~R. Rodrigues, I.~Oliveira, R.~Cunha, and M.~Netto, ``Deepdownscale: a deep
  learning strategy for high-resolution weather forecast,'' in \emph{2018 IEEE
  14th International Conference on e-Science (e-Science)}.\hskip 1em plus 0.5em
  minus 0.4em\relax IEEE, 2018, pp. 415--422.

\bibitem{mehrkanoon2019deep}
S.~Mehrkanoon, ``Deep shared representation learning for weather elements
  forecasting,'' \emph{Knowledge-Based Systems}, vol. 179, pp. 120--128, 2019.

\bibitem{krizhevsky2012imagenet}
A.~Krizhevsky, I.~Sutskever, and G.~E. Hinton, ``Imagenet classification with
  deep convolutional neural networks,'' in \emph{Advances in neural information
  processing systems}, 2012, pp. 1097--1105.

\bibitem{mehrkanoon2019deepneural}
S.~Mehrkanoon, ``Deep neural-kernel blocks,'' \emph{Neural Networks}, vol. 116,
  pp. 46--55, 2019.

\bibitem{salman2015weather}
A.~G. Salman, B.~Kanigoro, and Y.~Heryadi, ``Weather forecasting using deep
  learning techniques,'' in \emph{2015 international conference on advanced
  computer science and information systems (ICACSIS)}.\hskip 1em plus 0.5em
  minus 0.4em\relax IEEE, 2015, pp. 281--285.

\bibitem{grover2015deep}
A.~Grover, A.~Kapoor, and E.~Horvitz, ``A deep hybrid model for weather
  forecasting,'' in \emph{Proceedings of the 21th ACM SIGKDD International
  Conference on Knowledge Discovery and Data Mining}, 2015, pp. 379--386.

\bibitem{cadenas2009short}
E.~Cadenas and W.~Rivera, ``Short term wind speed forecasting in la venta,
  oaxaca, m{\'e}xico, using artificial neural networks,'' \emph{Renewable
  Energy}, vol.~34, no.~1, pp. 274--278, 2009.

\bibitem{monfared2009new}
M.~Monfared, H.~Rastegar, and H.~M. Kojabadi, ``A new strategy for wind speed
  forecasting using artificial intelligent methods,'' \emph{Renewable energy},
  vol.~34, no.~3, pp. 845--848, 2009.

\bibitem{rohrig2006application}
K.~Rohrig and B.~Lange, ``Application of wind power prediction tools for power
  system operations,'' in \emph{2006 IEEE Power Engineering Society General
  Meeting}.\hskip 1em plus 0.5em minus 0.4em\relax IEEE, 2006, pp. 5--pp.

\bibitem{liang2019deep}
X.~Liang, Y.~Zhang, G.~Wang, and S.~Xu, ``A deep learning model for
  transportation mode detection based on smartphone sensing data,'' \emph{IEEE
  Transactions on Intelligent Transportation Systems}, 2019.

\bibitem{hochreiter1997long}
S.~Hochreiter and J.~Schmidhuber, ``Long short-term memory,'' \emph{Neural
  computation}, vol.~9, no.~8, pp. 1735--1780, 1997.

\bibitem{xingjian2015convolutional}
S.~Xingjian, Z.~Chen, H.~Wang, D.-Y. Yeung, W.-K. Wong, and W.-c. Woo,
  ``Convolutional lstm network: A machine learning approach for precipitation
  nowcasting,'' in \emph{Advances in neural information processing systems},
  2015, pp. 802--810.

\bibitem{wang2017predrnn}
Y.~Wang, M.~Long, J.~Wang, Z.~Gao, and S.~Y. Philip, ``Predrnn: Recurrent
  neural networks for predictive learning using spatiotemporal lstms,'' in
  \emph{Advances in Neural Information Processing Systems}, 2017, pp. 879--888.

\bibitem{ronneberger2015u}
O.~Ronneberger, P.~Fischer, and T.~Brox, ``U-net: Convolutional networks for
  biomedical image segmentation,'' in \emph{International Conference on Medical
  image computing and computer-assisted intervention}.\hskip 1em plus 0.5em
  minus 0.4em\relax Springer, 2015, pp. 234--241.

\bibitem{chollet2017xception}
F.~Chollet, ``Xception: Deep learning with depthwise separable convolutions,''
  in \emph{Proceedings of the IEEE conference on computer vision and pattern
  recognition}, 2017, pp. 1251--1258.

\bibitem{lawhern2018eegnet}
V.~J. Lawhern, A.~J. Solon, N.~R. Waytowich, S.~M. Gordon, C.~P. Hung, and
  B.~J. Lance, ``Eegnet: a compact convolutional neural network for eeg-based
  brain--computer interfaces,'' \emph{Journal of neural engineering}, vol.~15,
  no.~5, p. 056013, 2018.

\bibitem{nair2010rectified}
V.~Nair and G.~E. Hinton, ``Rectified linear units improve restricted boltzmann
  machines,'' in \emph{Proceedings of the 27th international conference on
  machine learning (ICML-10)}, 2010, pp. 807--814.

\bibitem{bello2019attention}
I.~Bello, B.~Zoph, A.~Vaswani, J.~Shlens, and Q.~V. Le, ``Attention augmented
  convolutional networks,'' in \emph{Proceedings of the IEEE International
  Conference on Computer Vision}, 2019, pp. 3286--3295.

\bibitem{kingma2014adam}
D.~P. Kingma and J.~Ba, ``Adam: A method for stochastic optimization,''
  \emph{arXiv preprint arXiv:1412.6980}, 2014.

\end{thebibliography}

\end{document}